\documentclass[sigconf]{acmart}

\usepackage{hyperref}
\usepackage{url}
\usepackage{multirow}
\usepackage{longtable}
\usepackage{rotating}
\usepackage{booktabs}
\usepackage{algorithmic}
\usepackage{textcomp}
\usepackage{xcolor}
\usepackage{pdfpages}
\usepackage{graphicx}
\usepackage{ragged2e}
\usepackage{array}
\usepackage{color}
\usepackage{balance}
\AtBeginDocument{%
  }

\copyrightyear{2025}
\acmYear{2025}
\setcopyright{cc}
\setcctype{by}
\acmConference[KDD '25]{Proceedings of the 31st ACM SIGKDD Conference on Knowledge Discovery and Data Mining V.2}{August 3--7, 2025}{Toronto, ON, Canada}
\acmBooktitle{Proceedings of the 31st ACM SIGKDD Conference on Knowledge Discovery and Data Mining V.2 (KDD '25), August 3--7, 2025, Toronto, ON, Canada}
\acmDOI{10.1145/3711896.3737046}
\acmISBN{979-8-4007-1454-2/2025/08}

\begin{document}

\title{Merlin: Multi-View Representation Learning for Robust Multivariate Time Series Forecasting with Unfixed Missing Rates}

\author{Chengqing Yu}
\orcid{0000-0001-8314-8251}
\author{Fei Wang}
\orcid{0000-0002-3282-0535}
\authornote{Fei Wang and Yongjun Xu are the corresponding authors.}
\affiliation{%
  \institution{Institute of Computing Technology, }
  \city{Chinese Academy of Sciences}
  \country{\\State Key Laboratory of AI Safety}
}
\affiliation{%
  \institution{University of Chinese Academy of Sciences,}
  \city{Beijing}
  \country{China}
}
\email{{yuchengqing22b, wangfei, 19b}@ict.ac.cn}

\author{Chuanguang Yang}
\orcid{0000-0001-5890-289X}
\author{Zezhi Shao}
\orcid{0000-0002-0815-2768}
\author{Tao Sun}
\orcid{0000-0003-1692-3574}
\author{Tangwen Qian}
\orcid{0000-0001-5694-3831}
\affiliation{%
  \institution{Institute of Computing Technology, \\Chinese Academy of Sciences}
  \city{State Key Laboratory of AI Safety}
  \country{Beijing, China}
}
\email{{yangchuanguang, shaozezhi, suntao, qiantangwen}@ict.ac.cn}

\author{Wei Wei}
\orcid{0000-0003-4488-0102}
\affiliation{%
  \institution{School of Computer Science and Technology, \\Huazhong University of Science and Technology,}
  \city{Wuhan}
  \country{China}
}
\email{weiw@hust.edu.cn}

\author{Zhulin An}
\orcid{0000-0002-7593-8293}
\author{Yongjun Xu}
\orcid{0000-0001-6647-0986}
\authornotemark[1]
\affiliation{%
  \institution{Institute of Computing Technology, }
  \city{Chinese Academy of Sciences}
  \country{\\State Key Laboratory of AI Safety}
}
\affiliation{%
  \institution{University of Chinese Academy of Sciences,}
  \city{Beijing}
  \country{China}
}
\email{{anzhulin, xyj}@ict.ac.cn}

\renewcommand{\shortauthors}{Chengqing Yu et al.}

\begin{abstract}
Multivariate Time Series Forecasting (MTSF) involves predicting future values of multiple interrelated time series. Recently, deep learning-based MTSF models have gained significant attention for their promising ability to mine semantics (global and local information) within MTS data. However, these models are pervasively susceptible to missing values caused by malfunctioning data collectors. These missing values not only disrupt the semantics of MTS, but their distribution also changes over time. Nevertheless, existing models lack robustness to such issues, leading to suboptimal forecasting performance. To this end, in this paper, we propose \textbf{\underline{M}}ulti-Vi\textbf{\underline{e}}w \textbf{\underline{R}}epresentation \textbf{\underline{L}}earn\textbf{\underline{in}}g (\textbf{Merlin}), which can help existing models achieve semantic alignment between incomplete observations with different missing rates and complete observations in MTS. Specifically, Merlin consists of two key modules: offline knowledge distillation and multi-view contrastive learning. The former utilizes a teacher model to guide a student model in mining semantics from incomplete observations, similar to those obtainable from complete observations. The latter improves the student model's robustness by learning from positive/negative data pairs constructed from incomplete observations with different missing rates, ensuring semantic alignment across different missing rates. Therefore, Merlin is capable of effectively enhancing the robustness of existing models against unfixed missing rates while preserving forecasting accuracy. Experiments on four real-world datasets demonstrate the superiority of Merlin.
\end{abstract}


\maketitle

\newcommand\kddavailabilityurl{https://doi.org/10.5281/zenodo.15508818}

\ifdefempty{\kddavailabilityurl}{}{
\begingroup\small\noindent\raggedright\textbf{KDD Availability Link:}\\
The source code of this paper has been made publicly available at \url{\kddavailabilityurl}.
\endgroup
}

\section{Introduction}

Multivariate Time Series Forecasting (MTSF) widely exists in transportation, environment, weather and other areas \citep{shao2025spatial, deng2024parsimony, wang2023ai, huang2025foundation}. It consists of multiple numerical sequences that change over time, which typically contain global information (such as periodicity and trends) and local information (such as detailed changes) along the temporal dimension \citep{tan2022new, wu2025k2vae}. These global and local information constitute the semantics of time series, and fully mining these semantics is crucial for achieving accurate forecasting \citep{dong2024simmtm, qiu2025tab}.

\begin{figure}
  \centering
  \includegraphics[width=0.8\linewidth]{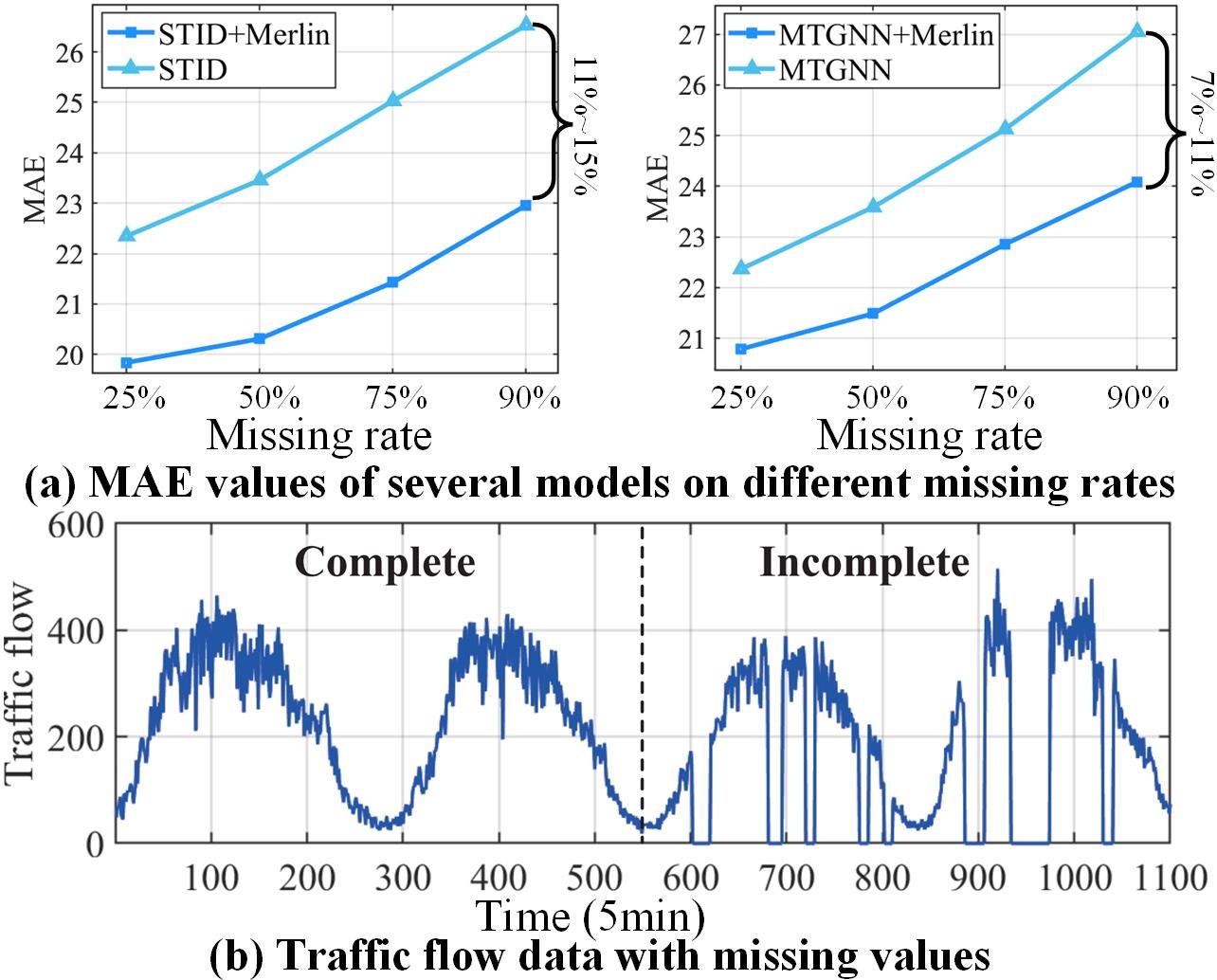}
  \caption{Examples of MTSF with missing values on PEMS04. (a) Although the forecasting errors of existing models increase significantly as the missing rate increases, Merlin can reduce their forecasting errors. (b) Missing values disrupt the global information and bring abnormal local information. The missing rates are unfixed across different time periods.}
  \label{fig1}
\end{figure}

Deep learning-based models, such as Spatial-Temporal Graph Neural Networks (STGNNs) \citep{shao2022decoupled, shao2022pre} and Transformers \citep{cheng2024multi, yu2024tfeformer}, can fully mine semantics from MTS, but they generally require complete observations to maintain accurate forecasting \citep{yu2025ginar+}. In reality, due to factors such as natural disasters and component failures, data collectors can easily malfunction and fail to output data normally \citep{zheng2023increase}. In this case, existing models only use incomplete observations to predict future values, limiting their performance \citep{cinifilling}. To illustrate, we evaluate the performance of several models \citep{wu2020connecting,RN859} under different missing rates on the PEMS04 dataset. As shown in Figure \ref{fig1}(a), the forecasting errors (Mean Absolute Error) of these models increase significantly as the missing rate increases.

To mitigate the adverse effects of incomplete MTS data, we delve deeper into a question: \textbf{how do missing values lead to the performance degradation of these models?} We find that a large number of missing values\footnote{Missing values in most datasets, such as PEMS04, are typically treated as zeros.} in historical observations can severely disrupt the semantics of MTS, while existing models exhibit poor robustness in this scenario. As shown in Figure \ref{fig1} (b), missing values disrupt the global information (such as periodicity) and introduce abnormal local information such as sudden changes (from normal to zero) and abnormal straight lines. Existing models easily capture these anomalies instead of the original semantics, leading to a degradation in performance. Besides, the distribution of missing values may change over time, leading to unfixed missing rates across different time periods. If existing models \citep{wu2020connecting, RN859} do not train separate models for each missing rate, their performance will significantly decline, exhibiting poor robustness to unfixed missing rates.

An intuitive solution to enhancing the performance of forecasting models is to use imputation methods to recover missing values and propose two-stage or end-to-end modeling approaches \citep{tran2023end, tang2020joint, yu2024ginar}. However, these methods face two challenges: (1) existing imputation methods \citep{miao2021generative, RN18, ahn2022comparison} usually require reconstructing both missing and normal values, which disrupts the local information and leads to error accumulation; (2) when dealing with unfixed missing rates, existing imputation methods \citep{du2023saits, zhou2023one, RN59} require training separate models for different missing rates to ensure data recovery accuracy. When imputation and forecasting models do not train separate models for each missing rate, their forecasting performance significantly declines, exhibiting poor robustness when dealing with unfixed missing rates.

To solve the above problems and achieve robust MTSF with unfixed missing rates, we believe that the key idea is to enhance the capability of forecasting models for semantic alignment, which includes \textbf{two objectives}: (1) complete observations have more accurate global and local information, and we need to help forecasting models mine similar semantics from incomplete observations; (2) different missing rates have varying impacts on the original semantics of MTS, and we need to help forecasting models align the semantics as much as possible between different missing rates.
To this end, we propose \textbf{\underline{M}}ulti-Vi\textbf{\underline{e}}w \textbf{\underline{R}}epresentation \textbf{\underline{L}}earn\textbf{\underline{in}}g (\textbf{Merlin}), taking advantage of offline knowledge distillation (KD) and multi-view contrastive learning (CL). 
\textbf{For objective 1}, based on the idea of offline knowledge distillation, we first train a teacher model using complete observations and then use the same architecture to obtain a student model, which is trained on incomplete observations. When training the student model, we input complete observations into the teacher model and transfer the generated representations and forecasting results as knowledge to constrain the student model. This ensures that the representations and forecasting results generated by the student model from incomplete observations are as similar as possible to those generated by the teacher model (the loss between them is as small as possible). In this way, the student model can mine semantics from incomplete observations that approximate those obtained from complete observations.
\textbf{For objective 2}, based on the idea of multi-view contrastive learning, we first treat incomplete observations from the same time point but with different missing rates as positive data pairs, and those from different time points as negative data pairs. Then, we propose a contrastive loss as a constraint for the student model to enhance the dissimilarity of negative data pairs and the similarity of positive data pairs. Through this approach, the student model can align the semantics between incomplete observations with different missing rates, without the need to train several separate models for different missing rates. As shown in Figure \ref{fig1} (a), Merlin can significantly enhance the robustness of existing models to unfixed missing rates and improve their forecasting performance.
\textbf{The main contributions can be outlined as follows:}
\begin{itemize}
    \item The existing models encounter robustness issues when dealing with unfixed missing rates, including two phenomena: (1) missing values significantly disrupt the semantics (global and local information) of MTS; (2) the missing rate typically changes over time, and existing models require training separate models for different missing rates.
    \item The key idea to achieving robust MTSF with unfixed missing rates is to help existing models achieve semantic alignment between incomplete observations with different missing rates and complete observations. To this end, we propose Merlin based on offline knowledge distillation and multi-view contrastive learning. 
    \item We design experiments on four real-world datasets. Results show that Merlin can enhance the performance of existing models more effectively than imputation methods. Besides, with the help of Merlin, existing models only need to be trained once to adapt to unfixed missing rates.
\end{itemize}

\section{Related Work}
\subsection{Time Series Analysis with Missing Values} \label{timesiss}

Existing works typically design imputation methods or end-to-end forecasting models to address the negative impact of missing values on MTSF \citep{yu2024ginar}. Since data collectors typically function properly during the early stages of deployment, and missing values occur gradually after a period of time, existing works generally assume that data collectors can collect complete data (complete observations) during the training phase \cite{cinifilling, marisca2024graph}. Under this assumption, in the training phase, existing works first mask complete observations to generate incomplete observations with different missing rates. Then, they use complete observations as labels and combine reconstruction loss to train imputation methods or end-to-end models \cite{chauhan2022multi, xu2023uncovering}. In the test phase, they only use incomplete observations to evaluate the performance of several models. 

Based on the above assumptions, in terms of both technology and scenario, end-to-end forecasting models \citep{tang2020joint, chen2023biased} first convert complete observations into incomplete observations based on random point missing. Then, they propose additional components (such as reconstruction loss or GNN) to directly model incomplete observations for each missing rate and achieve forecasting. Imputation methods \citep{RN59, chen2024laplacian, du2023saits} also train separate models for each missing rate to restore incomplete observations, using reconstruction loss and data recovery components (such as variant attention or GNN). 

In summary, mainstream works require training separate models for each missing rate, assuming that the data is complete during the training phase, and are typically conducted under the random point missing scenario. To be fair, we primarily conduct experiments and evaluate baselines based on the above scenario, assumptions, and settings. However, considering more realistic scenarios, we analyze other data missing scenarios in Section \ref{unfix} and evaluate the performance of existing works when the training sets contain missing values in Appendix \ref{teacher}.

\subsection{Time Series Forecasting Models}
Classic STGNNs \citep{liu2021new, RN58, RN56, deng2024disentangling} combine the Graph Convolutional Network (GCN) and sequence models to realize MTSF. At present, existing state-of-the-art (SOTA) STGNNs \citep{yi2023fouriergnn, li2021spatial} introduce graph learning to further improve their forecasting performance. Different from STGNNs, Transformer-based models \citep{wu2023interpretable, zhang2022crossformer, yu2023dsformer} combine temporal attention and spatial attention, or their variants, to improve forecasting results. However, these complex models suffer from high complexity and limited scalability \citep{YU2024102607, chen2023multi}. Currently, lightweight models based on Multi-Layer Perceptron (MLP) have gained widespread recognition. Chen et al. \citep{chen2023tsmixer} propose TSMixer, which uses all-MLP architecture to realize MTSF. Shao et al. \citep{RN859} analyze the core of modeling MTS and propose an MLP framework based on the Spatial-Temporal Identity (STID). In summary, a suitable MLP framework can achieve satisfactory results more efficiently than complex models. Considering that STID analyzes the characteristics of MTSF and performs satisfactorily on most datasets, it is selected as the backbone in this paper. In addition to STID, we also evaluate the performance improvement of Merlin on other backbones (STGNNs and Transformer) in Section \ref{trans}.

\subsection{Knowledge distillation}

Knowledge distillation can transfer valuable knowledge from the teacher model to the student model to improve the student model's performance \citep{xu2022contrastive}. Mainstream techniques include offline knowledge distillation and online knowledge distillation. Among them, offline knowledge distillation offers advantages such as good stability, high flexibility, and a simplified training process. It improves the ability of the student model by continually guiding it to align with the teacher model  \citep{wang2021mulde, yang2022cross}. Huang et al. \citep{huang2022feature} use knowledge distillation to help the student model achieve semantic alignment between low-resolution and high-resolution data. Even when the input size is reduced by 50\%, the student model can still maintain good performance. Monti et al. \citep{monti2022many} propose a trajectory forecasting model based on knowledge distillation and spatial-temporal Transformer, enabling the student model to perform well with only 25\% of historical observations. In summary, knowledge distillation helps the student model achieve satisfactory forecasting results when the effective information in the inputs is significantly reduced, thereby enhancing its robustness to missing values. 

\subsection{Contrastive learning}

Multi-view contrastive learning enhances the model's ability to align the semantics of similar samples across different views \citep{deng2022multi, deng2024multi, deng2021pulse, liang2024survey}. Woo et al. \citep{woo2021cost} treat the seasonal and trend components of MTS as different views and use contrastive learning to align their semantics. Yue et al. \citep{yue2022ts2vec} propose hierarchical contrastive learning to help the model improve their ability to align the semantics of MTS with different scales. 
Dong et al. \citep{dong2024simmtm} combine different masking ways with contrastive learning to realize time series forecasting. Results show that contrastive learning can align the semantics of different masked time series and enhance the forecasting effect. Overall, contrastive learning can enhance the model's ability to distinguish negative data pairs and align the semantics between positive data pairs \citep{liu2023timesurl}. Therefore, if we effectively construct positive data pairs, contrastive learning can align the semantics of incomplete observations with different missing rates.

\section{Methodology}

\subsection{Preliminaries}



\textbf{Multivariate time series forecasting} \citep{chengqing2023multi}. Given a historical observation tensor $X\in \mathbb{R}^{N_v \times N_H \times N_c}$ from $N_H$ time slices in history, the model can predict the value $Y\in \mathbb{R}^{N_v \times N_L}$ of the nearest. The core goal of MTSF is to construct a mapping function between the input $X\in \mathbb{R}^{N_v \times N_H \times N_c}$ and output $Y\in \mathbb{R}^{N_v \times N_L}$. $N_v$ is the number of sequences. $N_H$ is the historical length. $N_L$ is the future length. $N_c$ is the number of features.

\textbf{Multivariate time series forecasting with missing values} \citep{sridevi2011imputation}. In this task, we need to mask $M\%$ point randomly from the historical observation tensor $X\in \mathbb{R}^{N_v \times N_H \times N_c}$. After the above processing, a new input feature $X_M\in \mathbb{R}^{N_v \times N_H \times N_c}$ is obtained. The core goal of this task is to construct a mapping function between the input $X_M\in \mathbb{R}^{N_v \times N_H\times N_c}$ and output $Y\in \mathbb{R}^{N_v \times N_L}$.

\textbf{Offline knowledge distillation} \citep{dong2023momentum,yang2023online}. It can transfer valuable knowledge from the teacher model to the student model as an additional loss function, thereby constraining the student model's modeling process and improving its performance. When training the student, the teacher model does not need to update its parameters, and the teacher model is not used during test.

\textbf{Contrastive learning} \citep{zhang2024acvae,yang2022mutual}. Its main goal is to introduce contrastive loss to help the model enhance the dissimilarity of negative data pairs and the similarity of positive data pairs, thereby fully extracting invariant information from positive data pairs. Therefore, it can effectively improve the model's generalization.

\subsection{Overall Framework}

\begin{figure*}
\centering
\includegraphics[width=0.9\linewidth]{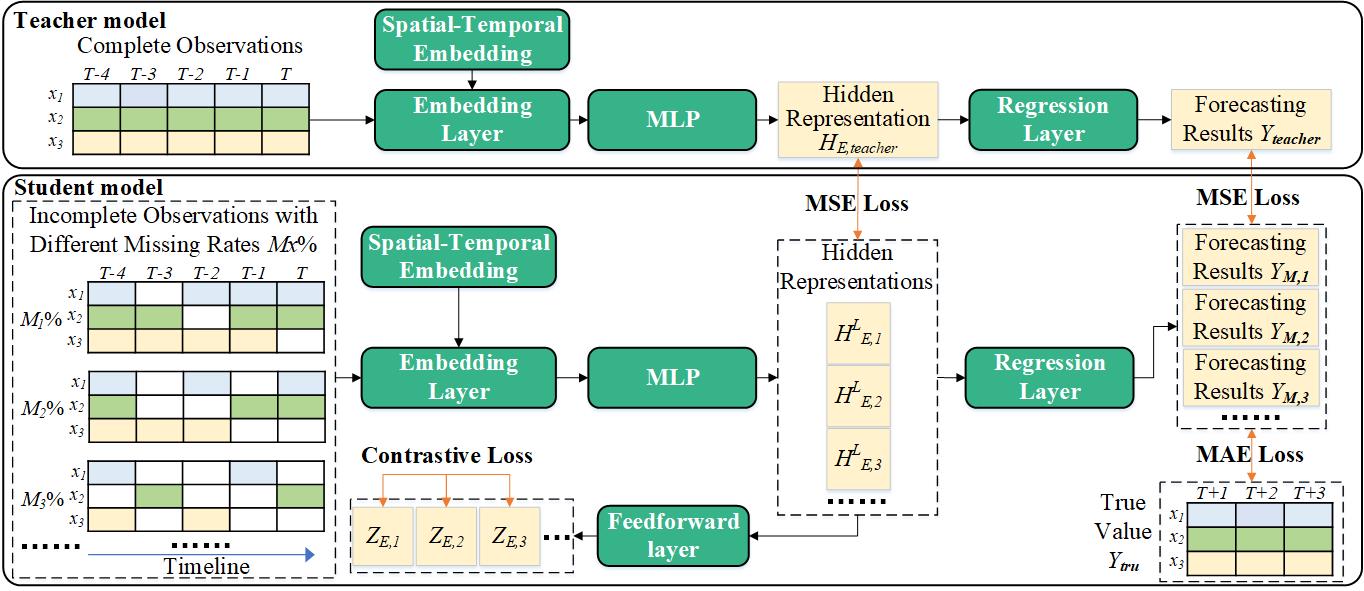}
\caption{The overall framework of our model. During the training phase, we first mask the raw data (complete observations) to obtain multiple new data with different missing rates (incomplete observations). The complete observations are used to train the teacher model, and the incomplete observations are used to train the student model. The teacher model guides the student model to align the semantics between complete and incomplete observations. Contrastive loss enhances the student model's robustness to incomplete observations with different missing rates. During the test phase, the trained student model is used to directly process incomplete observations and obtain forecasting results, without the use of the teacher model.}
\label{fig2}
\end{figure*}

The overall framework is shown in Figure \ref{fig2}. We use STID as the backbone and propose Merlin, which combines offline knowledge distillation with multi-view contrastive learning and is used to train STID. A basic introduction to STID is provided in Section \ref{backbone}. 

During the training phase, we follow the assumption of existing works that the training sets are complete, as discussed in Section \ref{timesiss}. Based on offline knowledge distillation, we first use complete observations to train the teacher model. Then, when using incomplete observations to train the student model, we transfer the representations and forecasting results generated by the teacher model as knowledge, providing constraints (MSE loss) for the student model. Meanwhile, for the representations obtained by the student model when encoding incomplete observations with different missing rates, we propose a contrastive loss based on multi-view contrastive learning to further constrain the student model to align the semantics between them. The MSE loss from knowledge distillation and the contrastive loss constitute the key components of Merlin, helping STID achieve semantic alignment between incomplete observations and complete observations. A detailed introduction to knowledge distillation and contrastive learning can be found in Section \ref{KD} and Section \ref{CL}, respectively.

During the test phase, the student model directly uses incomplete observations with different missing rates for forecasting, without the need for the teacher model. Compared to existing works, the use of data in both the training and test phases remains fair and reasonable. Although most related works assume that training sets are complete, training sets in the real world may contain missing values, which affects the performance of the teacher or imputation models. Therefore, in Appendix \ref{teacher}, we analyze the performance of Merlin when the training sets contain missing values.

\subsection{Backbone} \label{backbone}
In this section, we briefly introduce the basic structure of the backbone (STID), which is composed of a embedding layer, $L$ fully connected (FC) layer and a regression layer. The detailed description and definition of STID can be found in the reference \citep{RN859}. The basic modeling process of STID is shown as follows:

Step I: First, the embedding layer based on a FC layer is used to transform the input feature $X$ into a hidden representation $H$:
\begin{equation}
H =\text{FC}(X),
\end{equation}
where, $\text{FC}(\cdot)$ is the fully connected layer.

Step II: Then, the spatial-temporal identity embeddings ($S_E$, $T_{E}^{D}$ and $T_{E}^{W}$) are passed to $H$ as additional inputs to the encoder, helping the model capture spatial-temporal correlations.
\begin{equation}
H_E = \text{Concat}(H,S_E,T_{E}^{D},T_{E}^{W}),
\end{equation}
where, $\text{Concat}(\cdot)$ means concatenate several tensors. Assuming $N_v$ time series and $N_H$ time slots in a day and $N_w$ = 7 days in a week. $S_E \in R^{N_v*D}$ is the spatial identity embedding. $T_{E}^{D} \in R^{N_H*D}$ and $T_{E}^{W} \in R^{N_w*D}$ are temporal embedding. $D$ is the embedding size.

Step III: The encoder based on $L$ layers of MLP with the residual connection is used to mine the above representation $Z$. The $l$-th MLP layer can be denoted as:
\begin{equation}
H_E^{l+1} = \text{FC}(\text{Relu}(\text{FC}(H_E^l))) + H_E^l,
\end{equation}
where, $\text{Relu}(\cdot)$ is the activation function.

Step IV: Finally, based on the hidden representation $H_E^L$, the regression layer is used to obtain the forecasting results $Y$.
\begin{equation}
Y = \text{FC}(H_E^L).
\end{equation}

Compared to complex models, STID has two main advantages: (1) It introduces spatial-temporal identity embeddings to provide the model with additional global information, effectively mitigating the damage of missing values. (2) Based on the above embeddings, STID can effectively model MTS with the computational complexity of $O(N_H)$. Next, we show how to use hidden representations $H_E^L$ and forecasting results $Y$ for offline knowledge distillation with multi-view contrastive learning.

\subsection{Offline Knowledge Distillation} \label{KD}

We use two STID models as the student model and the teacher model. The inputs to the teacher model are complete observations $X$. It produces hidden representations $H_{E,Teacher}^L$ and forecasting results $Y_{Teacher}$. The inputs to the student model are incomplete observations $X_{M,1}$ to $X_{M,m}$. $m$ stands for the number of missing rates. It produces $m$ hidden representations $H_{E,1}^L$ to $H_{E,m}^L$ and $m$ forecasting results $Y_{M,1}$ to $Y_{M,m}$. The offline knowledge distillation consists of two components: the hidden representation distillation and the forecasting result distillation. 

\textbf{Hidden representation distillation:} It refers to transferring the representations produced by the teacher model to the student model, aiming to minimize the mean squared error (MSE) between the representations produced by the student model and those produced by the teacher model. Its specific formula is shown as follows:
\begin{equation}
L_{HD} = \sum_{i=1}^{m}\text{Mean}((H_{E,Teacher}^L-H_{E,i}^L)^2),
\end{equation}
where, $\text{Mean}(\cdot)$ is the mean of the Tensor.

\textbf{Forecasting result distillation:} It refers to transferring the forecasting results produced by the teacher model to the student model, with the objective of minimizing the MSE between the forecasting results produced by the student model and those produced by the teacher model. The specific formula is shown as follows:
\begin{equation}
L_{RD} = \sum_{i=1}^{m}\text{Mean}((Y_{Teacher}-Y_{M,i})^2).
\end{equation}

Based on $L_{HD}$ and $L_{RD}$, the teacher model can guide the student model to use incomplete observations to generate representations and forecasting results similar to those generated by the teacher model. In this way, the student model can achieve semantic alignment between incomplete observations and complete observations.

\subsection{Multi-View Contrastive Learning} \label{CL}

The missing rates are usually unfixed at different time periods. To further enhance the robustness of the student model and realize the semantic alignment between incomplete observations with different missing rates, we propose multi-view contrastive learning. We use incomplete observations with different missing rates at the same time point as positive data pairs, and use incomplete observations at different time point (other samples within a batch) as negative data pairs. For representations $H_{E,1}^L$ to $H_{E,m}^L$ encoded by incomplete observations with different missing rates, we achieve multi-view contrastive learning using the pairwise contrastive learning method. The specific steps are shown as follows:

Step I: Considering that appropriate dimension reduction can enhance the effectiveness of contrastive learning, we use an FC layer to reduce the dimension of the hidden representations $H_{E,1}^L$ to $H_{E,m}^L$, and obtain the representations $Z_{E,1}$ to $Z_{E,m}$. 
\begin{equation}
Z_{E,1} = \text{FC}(H_{E,1}^L).
\end{equation}

Step II: We use the $Z_{E,1}$ and $Z_{E,2}$ to obtain $2N_s$ samples. In $Z_{E,1}$ and $Z_{E,2}$, the corresponding two samples form a positive data pair, while the other samples are their negative data pairs. Based on cross-entropy loss, the contrast loss between any two samples $z_{E,i}$ and $z_{E,j}$ is calculated using the following formula:
\begin{equation}
l_{i, j} =-\text{log}(\frac{\text{exp}(\text{sim}(z_{E,i},z_{E,j})/\tau)}{\sum_{k=1 \& k \neq i }^{2N_s}\text{exp}(\text{sim}(z_{E,i},z_{E,k})/\tau)}),
\end{equation}
where, $\text{exp}(\cdot)$ is the exp function. $\text{sim}(\cdot)$ is the Cosine similarity. $N_s$ is the number of samples. $\tau$ is the temperature parameter.

Step III: Based on the above formula, we calculate the contrastive loss for all samples between $Z_{E,1}$ and $Z_{E,2}$, and obtain $L_{Z1, Z2}$.
\begin{equation}
L_{Z1, Z2} = \frac{1}{2N_s}\sum_{k=1}^{N_s}(l_{2k-1, 2k}+l_{2k, 2k-1}).
\end{equation}

Step IV: The final multi-view contrastive learning loss $L_{CL}$ is obtained by computing the pairwise contrastive losses from $Z_{E,1}$ to $Z_{E,m}$.
\begin{equation}
L_{CL} = \frac{2}{m(m-1)}(\sum_{Zj=Zi}^{m}\sum_{Zi=1}^{m-1}L_{Zi, Zj}).
\end{equation}

\subsection{Loss Function}

Since MTSF is a regression task, we also need to carry out the supervised learning process. To this end, we also incorporate the ground truth and L1 loss \citep{challu2023nhits} to train the student model.
\begin{equation}
L_{Pre} = \sum_{i=1}^{m}\text{Mean}(|Y_{tru}-Y_{M,i}|,
\end{equation}
where, $Y_{tru}$ is the ground truth. $|\cdot|$ stands for absolute value.

Finally, we need to effectively combine all the above loss functions. In this paper, we use the method of adding all loss functions, which can avoid the problem of information forgetting \citep{gou2023hierarchical}. The final loss $L_{Finally}$ function is shown as follows:

\begin{equation}
L_{Finally} = \beta_{1} * L_{Pre} + \beta_{2}*(L_{HD} + L_{RD}) + \beta_{3}*L_{CL},
\end{equation}
where, $\beta$ stands for the weight of the loss. $\beta_2$ and $\beta_3$ need to be divided by the current epoch size in each epoch. After completing the process of the training phase, the test phase is performed by using only the student model, without using the teacher model.

\section{Experiment and Analysis}
\subsection{Experimental Design}
\textbf{Datasets.} We select four real-world datasets from different domains: METR-LA\footnotemark[2], PEMS04\footnotemark[3], China AQI\footnotemark[4], and Global Wind\footnotemark[5]. The basic statistics for these datasets are shown in Table \ref{tab2}.

\begin{table}
\centering
  \caption{The statistics of four datasets.}
  \label{tab2}
  \scalebox{1}{\begin{tabular}{cccc}
\toprule
    Datasets&Variates &Timesteps&Granularity\\
\midrule
METR-LA& 207&34272&5 minutes\\
PEMS04& 307&16992&5 minutes\\
China AQI&1300&41506&1 hour\\
Global Wind&2908&10957&1 day\\
\toprule
\end{tabular}}
\end{table}

\footnotetext[2]{https://github.com/liyaguang/DCRNN}

\footnotetext[3]{https://github.com/guoshnBJTU/ASTGNN/tree/main/data}

\footnotetext[4]{https://quotsoft.net/air/}

\footnotetext[5]{https://www.ncei.noaa.gov/}

\textbf{Baselines.} We select baselines from three perspectives:
(1) Three SOTA one-stage forecasting models: TSMixer \citep{chen2023tsmixer},  FourierGNN \citep{yi2023fouriergnn}, and DSformer \citep{yu2023dsformer}.
(2) To demonstrate the improvement of Merlin on STID, we compare the STID+Merlin with the raw STID. Besides, we combine four imputation methods with STID to obtain multiple two-stage models: STID+GATGPT \citep{chen2023gatgpt}, STID+SPIN \citep{RN59}, STID+GPT2 \citep{zhou2023one} and STID+TI-MAE \citep{li2023ti}. 
(3) We combine existing multivariate time series forecasting models with imputation models, and obtain several two-stage models as baselines: iTransformer \citep{liu2023itransformer} + TimesNet \citep{RN18}, MTGNN \citep{wu2020connecting} + SPIN, TimeMixer\citep{wang2023timemixer} + GATGPT, and DUET\citep{qiu2025duet} + GPT2. The previous method for each combination is the forecasting model. 

\textbf{Setting.} 
We design the experiments from the following aspects:
(1) According to ratios in \citep{shao2023exploring, qiu2024tfb}, four datasets are uniformly divided into training sets, validation sets, and testing sets.
(2) The history length and future length are 12. All metrics are calculated as the average of the 12-step forecasting results. 
(3) We randomly assign mask points with ratios of 25\%, 50\%, 75\%, and 90\%. The value of the masked point is uniformly set to zero. Experiments are repeated with 5 different random seeds for each model. The final metrics are calculated as the mean value of repeated experiments. 
(4) To prove the robustness of Merlin, we train it once, using data with multiple missing rates. Specifically, the student model is trained simultaneously using data with missing rates of 25\%, 50\%, 75\%, and 90\%. 
(5) Our code is available at this link\footnotemark[6].

\footnotetext[6]{https://github.com/ChengqingYu/Merlin}

\textbf{Metrics.} 
Three classical metrics are used, including MAE (Mean Absolute Error), RMSE (Root Mean Square Error) and MAPE (Mean Absolute Percentage Error) \citep{liu2020new}. The smaller these metrics are, the lower the forecasting errors. 

\subsection{Main Results}\label{main}

\begin{table*}
\centering
\caption{Performance comparison results of several models. The best results are shown in \textbf{bold}. }\label{tab4}
\scalebox{0.85}{\begin{tabular}{cccccccccccccc}
\toprule
\multirow{2}*{Datasets} & \multirow{2}*{Models}&\multicolumn{3}{c}{Missing rate 25\%} &\multicolumn{3}{c}{Missing rate 50\%} &\multicolumn{3}{c}{Missing rate 75\%} &\multicolumn{3}{c}{Missing rate 90\%}\\
\cline{3-14} 
& & MAE & MAPE & RMSE & MAE & MAPE & RMSE & MAE & MAPE & RMSE & MAE & MAPE & RMSE \\

\midrule

\multirow{13}*{METR-LA} 

&DSformer&3.44&10.45&7.21 & 3.61& 11.15&7.52&3.82& 12.31&  8.11& 4.05 & 13.57& 8.84\\
&FourierGNN &3.42&10.39&7.17 &3.59&10.96&7.47&3.78&12.18&7.95&3.97&12.89&8.62	\\
&TSMixer &3.45&10.48&7.25 &3.60&11.09&7.53&3.84&12.34&8.21&4.01&13.42&8.76	\\
&iTransformer+TimesNet &3.37&10.31& 7.02&3.51&10.77&7.39&3.72&11.61&7.81&3.89&12.47&8.35\\
&MTGNN+SPIN &3.26&9.79&6.73 &3.45&10.45&7.18&3.59&10.97&7.56&3.84&12.31&8.24\\
&TimeMixer+GATGPT &3.35&9.97&6.91 &3.49&10.65&7.27&3.63&11.23&7.59&3.86&12.29&8.28\\
&DUET+GPT2  &3.32&9.95&6.87 &3.48&10.62&7.26&3.61&11.18&7.54&3.87&12.42&8.31\\
&STID (Raw) &3.39&10.27&7.05 & 3.53&10.88& 7.45& 3.75&11.83&7.86&3.92&12.59&8.43\\
&STID+SPIN &3.28&9.84& 6.79 & 3.44& 10.39& 7.20&3.60&11.06&7.55&3.78&12.02&7.93\\
&STID+GPT2  &3.31&9.91&6.83 & 3.43& 10.44& 7.18&3.55&10.85&7.49&3.75&11.84&7.86\\
&STID+TI-MAE   &3.34&9.92&6.88 & 3.46& 10.51& 7.22&3.58&10.93&7.52&3.76&11.85&7.89\\
&STID+GATGPT   &3.27&9.85&6.75 & 3.39& 10.35& 7.06&3.52&10.87&7.43&3.73&11.76&7.85\\
&STID+Merlin &\textbf{3.21}	&\textbf{9.41}	&\textbf{6.64}&	\textbf{3.28}	&\textbf{9.85}	&\textbf{6.78}	&\textbf{3.43}	&\textbf{10.42}	&\textbf{7.18}	&\textbf{3.65}	&\textbf{11.25}	&\textbf{7.64}\\

\midrule

\multirow{13}*{PEMS04} 

&DSformer&23.65&16.24&38.34&	24.37&	16.29&	39.83&	25.87&	17.66&	41.73	&27.43	&18.41	&43.17\\
&FourierGNN&22.95&15.42&37.06&	24.06&	16.15	&39.48&	25.69	&17.49&	41.48&	27.29	&18.32	&42.81\\
&TSMixer&23.37&15.81&37.40&	24.26&	16.04	&39.57&	25.76	&17.54&	41.65&	28.04&	18.58&	43.49\\
&iTransformer+TimesNet &21.64&15.08&34.56&	22.75&	15.34&36.57&	24.18&	16.06&	39.42&	25.57&17.56&41.23\\
&MTGNN+SPIN&20.94&14.71&33.52&	21.78	&15.07	&35.22	&23.36	&15.78	&37.64	&24.92	&16.63	&40.77\\
&TimeMixer+GATGPT&21.38&14.87&34.42&	22.42	&15.13	&36.03	&23.98	&15.94	&39.23	&25.43	&17.56	&41.10\\
&DUET+GPT2&21.29&14.84&34.37	&22.51	&15.21	&36.23	&24.16	&16.02	&39.54	&25.25	&17.34	&40.89\\
&STID (Raw) &22.35&15.08&36.05&	23.46	&16.18	&37.92 &25.03	&17.13&	40.87	&26.53&	17.91&	42.32	\\
&STID+SPIN&20.47&14.22&32.64&21.69	&15.02	&34.97	&23.26	&15.57	&37.29	&24.79&16.52	&40.58\\
&STID+GPT2&20.76	&14.36&32.96&21.63&14.92	&34.67	&23.15	&15.51	&37.21	&24.51	&16.37	&40.16\\
&STID+TI-MAE	&20.81&14.52&33.27&21.75	&15.05	&35.04&23.04	&15.42	&37.15	&24.34	&16.32	&39.77\\
&STID+GATGPT&20.42&14.15&32.58&21.36	&14.83	&34.12	&22.87	&15.26	&36.94	&24.09	&15.94	&39.36\\
&STID+Merlin&\textbf{19.84}	&\textbf{13.85}&\textbf{31.65}&\textbf{20.31}	&\textbf{14.07}	&\textbf{32.37}	&\textbf{21.43}	&\textbf{14.95}	&\textbf{34.45}	&\textbf{22.96}	&\textbf{15.56}	&\textbf{37.07}\\

\midrule

\multirow{13}*{China AQI} 

&DSformer               &17.52&35.36&29.81   &19.32	&39.27&31.74	  &20.47&43.66&33.85	&23.48	&45.52	&38.69\\
&FourierGNN	            &17.35&35.25&29.27   &19.14	&38.43&31.29      &20.96&43.83&34.04	&24.06	&46.25	&39.43\\
&TSMixer	            &17.03&34.83&28.84   &18.85	&37.68&30.62	  &20.28&43.54&33.42	&23.65	&45.93	&38.94\\
&iTransformer+TimesNet	&16.49&32.06&28.07   &17.79	&35.76&29.45	  &20.03&42.76&32.98	&23.04	&45.65	&37.68\\
&MTGNN+SPIN	            &15.83&31.89&27.21   &16.97	&34.13&28.79	  &18.91&40.56&30.77	&22.18	&44.92	&36.03\\
&TimeMixer+GATGPT	    &16.09&32.56&27.48   &17.39	&34.82&29.04	  &19.86&41.37&32.27	&22.57	&45.36	&36.81\\
&DUET+GPT2	            &15.97&32.04&27.35   &17.04	&34.73&28.91      &19.34&40.85&31.84	&22.30	&45.01	&36.39\\

&STID (Raw)	    &16.86&34.46&28.81   &18.56	&39.95	&30.24	 &20.36	&43.63	&33.17	 &23.42	&45.84	&38.54\\
&STID+SPIN	    &15.39&30.25&26.64   &16.83	&34.67	&28.69	 &18.43	&40.43	&30.06	 &21.86	&44.78	&35.83\\
&STID+GPT2	    &15.53&30.89&26.92   &16.91	&34.88	&28.75	 &18.35	&40.26	&29.97	 &21.45	&44.26	&34.96\\
&STID+TI-MAE	&15.74&31.72&27.09   &17.08	&35.24	&28.97	 &18.29	&40.07	&29.82	 &21.58	&44.53	&35.23\\
&STID+GATGPT	&15.46&30.53&26.87   &16.67	&34.34	&28.43	 &18.14	&39.96	&29.65	 &21.07	&43.79	&34.36\\

&STID+Merlin	&\textbf{15.21}&\textbf{30.42}&\textbf{26.38}&\textbf{15.84}&\textbf{32.06}	&\textbf{27.25}	&\textbf{17.43}	&\textbf{35.68}	&\textbf{29.34}	&\textbf{20.18}	&\textbf{43.36}	&\textbf{33.31}\\

\midrule

\multirow{13}*{Global Wind} 

&DSformer	&5.92&35.78&8.42&6.09	&36.04	&8.65	&6.17	&36.28	&8.77	&6.31	&36.52	&8.97\\
&FourierGNN	&5.95&35.81&8.43&6.11	&36.12	&8.69	&6.18	&36.33	&8.79	&6.36	&36.47	&9.08\\
&TSMixer	&5.93&35.75&8.39&6.07	&35.94	&8.62	&6.21	&36.12	&8.84	&6.33	&36.34	&9.02\\
&iTransformer+TimesNet	&5.88&35.66&8.30&6.01	&35.77	&8.55	&6.12	&35.95	&8.74	&6.23	&36.08	&8.86\\
&MTGNN+SPIN	&5.84&35.18&8.23&5.94	&35.42	&8.45	&6.10	&35.69	&8.69	&6.18	&36.04	&8.77\\
&TimeMixer+GATGPT	&5.86&35.27&8.25&5.99	&35.53	&8.54	&6.11	&35.83	&8.71	&6.21	&36.25	&8.83\\
&DUET+GPT2	&5.85&35.38&8.29&5.95	&35.54	&8.52	&6.09	&35.72	&8.65	&6.17	&35.94	&8.75\\
&STID (Raw)	&5.91&35.53&8.38&6.05	&35.62	&8.57	&6.14	&35.87	&8.73	&6.25	&36.31	&8.89\\
&STID+SPIN	&5.80&34.57&8.25&5.94	&35.18	&8.41	&6.12	&35.41	&8.69	&6.21	&36.27	&8.82\\
&STID+GPT2	&5.83&35.01&8.27&5.93	&35.09	&8.39	&6.08	&35.34	&8.60	&6.19	&35.45	&8.83\\
&STID+TI-MAE	&5.84&35.06&8.28&5.94	&35.15	&8.43	&6.09	&35.26	&8.62	&6.18	&35.41	&8.76\\
&STID+GATGPT	&5.82&34.75&8.26&5.91	&34.92	&8.37	&6.07	&35.08	&8.57	&6.15	&35.34	&8.73\\
&STID+Merlin	&\textbf{5.76}	&\textbf{34.28}	&\textbf{8.17}&\textbf{5.85}	&\textbf{34.59}	&\textbf{8.32}	&\textbf{5.97}	&\textbf{34.83}	&\textbf{8.49}	&\textbf{6.08}	&\textbf{35.04}	&\textbf{8.64}\\

\toprule

\end{tabular}}
\end{table*}

Table \ref{tab4} shows the performance comparison results of all baselines and Merlin. We can draw the following conclusions: 
(1) Compared with two-stage models, the forecasting errors of all single-stage models are larger. The main reason is that existing single-stage models are not robust to incomplete observations.
(2) Compared with imputation methods, Merlin improves the forecasting performance of STID more effectively. The main reason is that Merlin effectively combines the advantages of multi-view contrastive learning and offline knowledge distillation, which can significantly enhance the robustness of STID in modeling incomplete observations.
(3) STID+Merlin works better than all baselines in all cases. Firstly, we select the high-performance STID as our backbone model, which introduces spatial-temporal embeddings to provide additional global information for the model, helping to mitigate the impact of missing values. Secondly, we introduce offline knowledge distillation to guide STID on how to align the semantics between incomplete observations and complete observations, thereby enhancing STID's robustness to incomplete observations. Finally, we propose multi-view contrastive learning to achieve semantic alignment between incomplete observations with different missing rates, further improving its robustness. Therefore, STID+Merlin can achieve the best forecasting results on all datasets and all missing rates. 
In the next section, we will further evaluate Merlin's performance improvement effects on other backbones.

\subsection{Transferability of Merlin}\label{trans}

To validate the effectiveness and transferability of Merlin, we choose three other models (TimeMixer, DUET, and MTGNN) as backbones and compare the performance of Merlin with other imputation methods (GATGPT, GPT2, TI-MAE and SPIN). Table \ref{tab5} shows the MAE values of Merlin and other imputation methods. We can draw the following conclusions:
(1) One-stage models struggle to perform well. Specifically, missing values make it difficult for existing models to fully mine semantics from incomplete observations, resulting in unsatisfactory results.
(2) Compared with imputation methods, Merlin can better restore the performance of all backbone models. The experimental results fully prove the transfer ability and practical value of Merlin, which can effectively enhance the robustness of all backbones.

\begin{table}
\centering
\caption{MAE values of Merlin and other methods (The best results are shown in \textbf{bold}).}\label{tab5}
\scalebox{0.75}{\begin{tabular}{cc|cccc|cccc}
\toprule

\multirow{2}*{Backbone} & \multirow{2}*{Methods}&\multicolumn{4}{c|}{METR-LA} &\multicolumn{4}{c}{Global Wind}\\

\cline {3-10}
& &25\% &50\% &75\% &90\%&25\% 	&50\% &75\% &90\%\\

\midrule
\multirow{6}*{\rotatebox{90}{DUET}}&+Merlin	
&\textbf{3.26}&\textbf{3.38} &\textbf{3.53} &\textbf{3.77}
&\textbf{5.82}&\textbf{5.92}	&\textbf{6.06}	&\textbf{6.13}\\
&+GATGPT
&3.28&3.45&3.56&3.83
&5.83&5.94&6.08&6.15\\
&+GPT2	
&3.32&3.48&3.61&3.87
&5.85&5.95	&6.09	&6.17\\
&+TI-MAE	
&3.35&3.49&3.59&3.86
&5.86&5.97&6.11&6.21\\
&+SPIN	
&3.29&3.51&3.64&3.91
&5.83&5.96&6.12&6.25\\
&raw	
&3.41&3.56&3.79&3.95
&5.93&6.07&6.15&6.29\\
\midrule 
\multirow{6}*{\rotatebox{90}{TimeMixer}}&+Merlin
&\textbf{3.31}&\textbf{3.42} &\textbf{3.57} &\textbf{3.81}
&\textbf{5.83}&\textbf{5.95} &\textbf{6.08} &\textbf{6.17}\\
&+GATGPT
&3.35&3.49&3.63&3.86
&5.86&5.99    &6.11  &6.21\\
&+GPT2	
&3.39&3.54&3.67&3.89
&5.87&6.02&6.13&6.23\\
&+TI-MAE	
&3.41&3.55&3.71&3.95
&5.91&6.03&6.12&6.24\\
&+SPIN	
&3.36&3.56&3.74&3.93
&5.85&6.05&6.14&6.27\\
&raw	
&3.43&3.59&3.81&3.99
&5.94&6.08&6.16&6.32\\
\midrule
\multirow{6}*{\rotatebox{90}{MTGNN}}&+Merlin
&\textbf{3.24}&\textbf{3.37} &\textbf{3.52} &\textbf{3.75}
&\textbf{5.81}&\textbf{5.93} &\textbf{6.04} &\textbf{6.11}\\
&+GATGPT
&3.27&3.42&3.57&3.79
&5.85&5.96&6.08&6.16\\
&+GPT2	
&3.31&3.43&3.61&3.85
&5.87&5.97&6.11&6.19\\
&+TI-MAE	
&3.33&3.46&3.64&3.87
&5.88&5.99&6.13&6.21\\
&+SPIN	
&3.26&3.45&3.59&3.84
&5.84&5.94&6.10&6.18\\
&raw	
&3.39&3.55&3.77&3.92
&5.92&6.06&6.17&6.28\\

\toprule

\end{tabular}}
\end{table}

\subsection{Ablation Experiments}

\begin{figure}
\centering
\includegraphics[width=\linewidth]{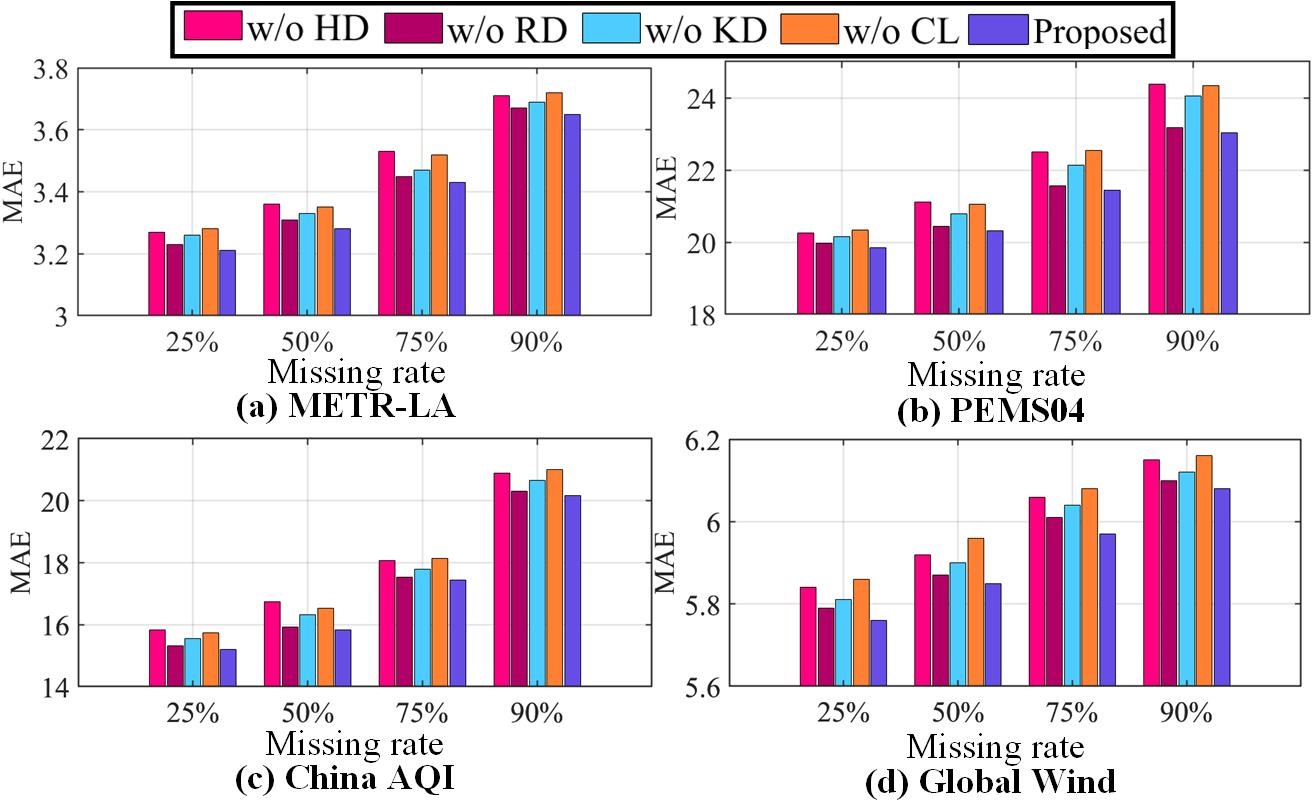}
\caption{Results of ablation experiments. w/o HD represents the removal of hidden representation distillation. w/o RD stands for the removal of forecasting result distillation. w/o KD indicates the deletion of the teacher model and the offline knowledge distillation. w/o CL represents the removal of multi-view contrastive learning.}
\label{fig4}
\end{figure}

We conduct ablation experiments from the following four perspectives:
(1) \textbf{w/o HD}: We remove the hidden representation distillation. 
(2) \textbf{w/o RD}: We remove the forecasting result distillation. 
(3) \textbf{w/o KD}: We removed the teacher model and knowledge distillation.In this case, STID achieves contrastive learning using both complete and incomplete observations.
(4) \textbf{w/o CL}: We remove the multi-view contrastive learning. 

Figure \ref{fig4} shows the results of the ablation experiment. We can draw the following conclusions:
(1) The forecasting result distillation has the least effect on the results. It shows that as long as the encoder can fully mine semantics from MTS, the decoder can realize effective forecasting.
(2) When either the hidden representation distillation or the contrastive learning is removed individually, the model performance drops significantly. This is mainly because both components play a critical role in achieving semantic alignment.
(3) When contrastive learning is used without knowledge distillation, the model's forecasting performance also declines, which further demonstrates the importance of knowledge distillation in aligning the semantics between complete and incomplete observations.

\subsection{Hyperparameter Analysis}\label{hyper}

\begin{figure}
\color{red}
\centering
\includegraphics[width=0.9\linewidth]{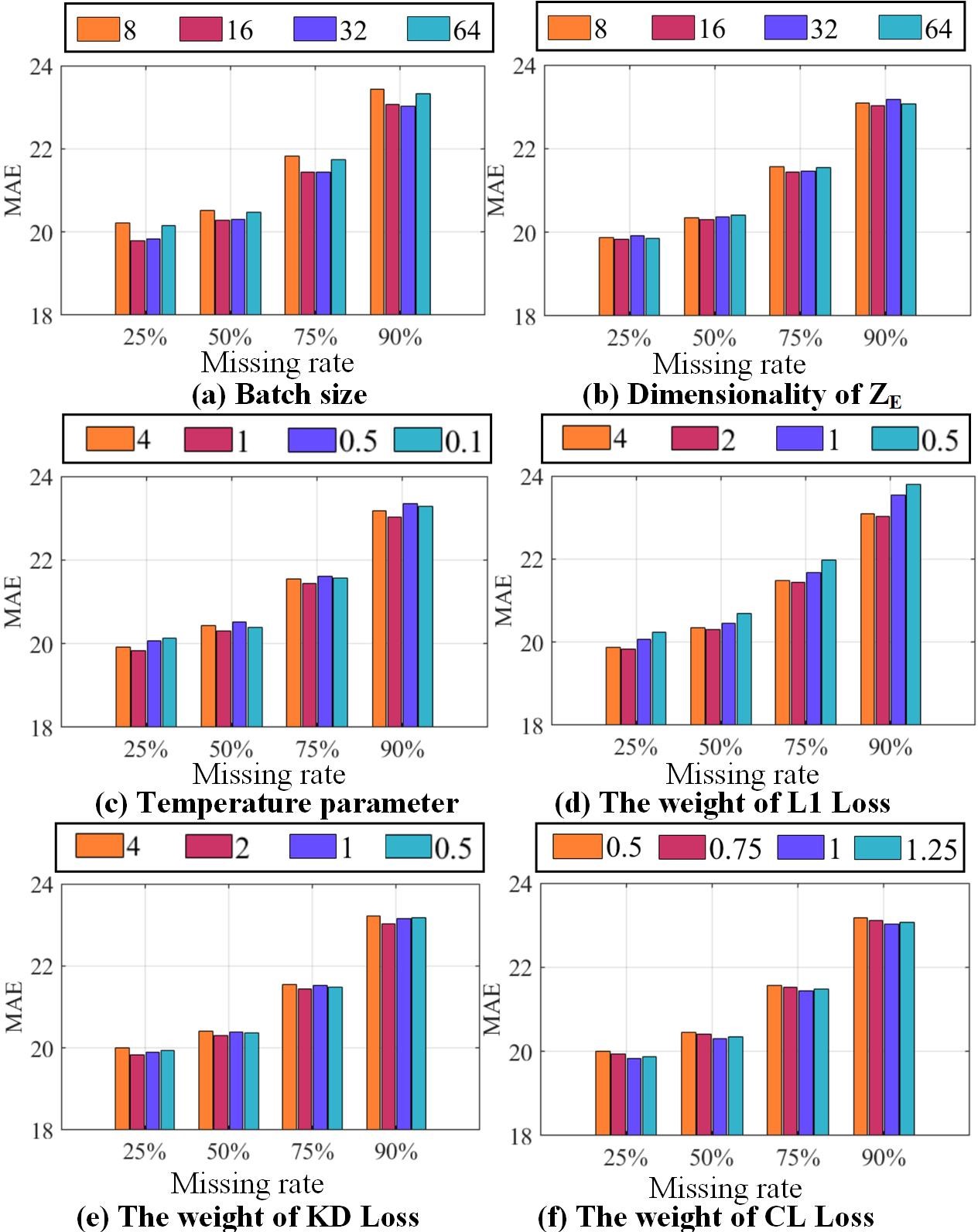}
\caption{Results of hyperparameter experiments (PEMS04).}
\label{fig9}
\end{figure}

Table \ref{tab3} presents the main hyperparameters of Merlin and STID. We evaluate six hyperparameters that have the most significant impact on Merlin's performance: Three weights of the loss ($\beta_1$, $\beta_2$ and $\beta_3$), batch size, temperature parameter, and the dimensionality of $Z_E$ \citep{chen2020simple}. The experimental results on PEMS04 are shown in Figure \ref{fig9}. We can draw the following conclusions:
(1) The dimensionality of $Z_E$ has limited influence on the experimental results, proving that contrastive learning is capable of achieving effective semantic alignment with relatively few parameters.
(2) Appropriately increasing the batch size can improve the forecasting accuracy of STID. A larger batch size increases the number of negative data pairs, thereby enhancing the model's robustness and its ability to capture semantics. However, the batch size that is too large can lead to underfitting issues.
(3) Proper balance of temperature parameter is important to improve the effect of contrastive learning. Appropriately reducing the temperature parameter can improve the performance of STID and enhance its convergence. However, setting the temperature parameter too small may lead to the problem of local optimality.
(4) Among the three weights of the loss, $\beta_1$ plays the most significant role. This is primarily because the knowledge distillation and contrastive loss are mainly used as constraints during the early training stages, whereas the model's performance in the later stages primarily depends on the L1 loss.

\begin{table}
\centering
  \caption{Values of the corresponding hyperparameters.}
  \label{tab3}
  \scalebox{0.85}{\begin{tabular}{ccc}
\toprule
    Methods&Config&Values\\
\midrule

\multirow{6}*{Merlin} 

& weight of L1 Loss $\beta_1$ &2\\
& weight of KD Loss $\beta_2$ &2\\
& weight of CL Loss $\beta_3$ &1\\
&temperature parameter&1\\
&dimensionality of $Z_E$&16\\
& batch size&16\\
\midrule
\multirow{14}*{STID} 
&optimizer&Adam \citep{kingma2014adam}\\
&learning rate&0.0002\\
& weight decay&0.0001\\
&embedding size&64\\
&node embedding size&64\\
&temporal embedding size (day)&64\\
&temporal embedding size (week)&64\\
&number of layers&3\\
&dropout&0.15\\
&learning rate schedule&MultiStepLR\\
&clip gradient normalization&5\\
&milestone&[1,10,25,50,75,90,100]\\
&gamme&0.5\\
&epoch&101\\

\toprule
\end{tabular}}
\end{table}

\subsection{Experiment on Multivariate Time Series Forecasting with Unfixed Missing Rates}\label{unfix}

To simulate the unfixed missing rates in real-world, we conduct the following experiments:
(1) For the training and validation sets, we additionally process the data into four forms with missing rates of 25\%, 50\%, 75\%, and 90\%.
(2) For the test set, we divide the time series into different segments based on time and randomly mask each segment with missing rates of 25\%, 50\%, 75\%, and 90\%.
(3) For STID+Merlin, we train it as described in this paper: the unmasked data is used to train the teacher model, while the masked data is used to train the student model. Only the student model is used on the test set.
(4) Baselines use three training ways: the first way involves training separate models for each missing rate, with the corresponding model selected for forecasting on the test set based on the current missing rate. The second way uses a single model trained on masked data with all four missing rates, which is directly tested on the test set. The final way trains a model using the unmasked data, which is directly tested on the test set.

Table \ref{tab10} shows the results of several models on unfixed missing rates. The following conclusions can be drawn: 
(1) Trained only once, STID+Merlin achieves best results across all datasets. Experimental results demonstrate that STID+Merlin can effectively handle the unfixed missing rates.
(2) For other baselines, training models separately for each missing rate performs better than training a single model for all missing rates, demonstrating that existing methods are limited in both practical value and robustness in MTSF with unfixed missing rates.
(3) If a forecasting model is trained using only complete data, its forecasting performance significantly declines when data missing occurs. This demonstrates the poor robustness of existing models in real-world scenarios.

\begin{table}
\centering
\caption{Performance comparison results of several models on unfixed missing rates.}\label{tab10}
\scalebox{0.85}{\begin{tabular}{ccccc}
\toprule
Datasets & Methods &MAE&MAPE&RMSE\\

\midrule

\multirow{12}*{\rotatebox{90}{METR-LA}}
&STID+Merlin&	\textbf{3.41}&	\textbf{10.27}&\textbf{7.08}	\\
&STID+GATGPT (Separately)& 3.48&	10.74&7.31	\\
&STID+GATGPT (Together)& 3.55&	10.87&7.48	\\
&STID (Separately)& 3.61&	11.12&7.59	\\
&STID (Together)& 3.73&	11.78&7.84	\\
&STID (Complete) & 4.06&13.62&8.91\\
\cline {2-5}
&MTGNN+Merlin& \textbf{3.47}&	\textbf{10.65}&\textbf{7.25}	\\
&MTGNN+GATGPT (Separately)& 3.53&	10.85&7.44	\\
&MTGNN+GATGPT (Together)& 3.58&	11.04&7.52	\\
&MTGNN (Separately) &3.64&	11.34&7.63	\\
&MTGNN (Together)&3.77& 12.15&	7.98	\\
&MTGNN (Complete) &4.09 &13.77&9.01\\
\midrule

\multirow{12}*{\rotatebox{90}{PEMS04}}
&STID+Merlin&	\textbf{21.18}&	\textbf{14.79}&\textbf{34.06}	\\
&STID+GATGPT (Separately)& 22.15&	15.34&35.58	\\
&STID+GATGPT (Together)& 24.03&	15.87&39.23	\\
&STID (Separately)& 24.52&	16.44&39.97	\\
&STID (Together)& 25.64&	17.66&41.43	\\
&STID (Complete) & 31.27&22.58&47.41\\
\cline {2-5}
&MTGNN+Merlin& \textbf{22.34}&	\textbf{15.08}&	\textbf{35.94}\\
&MTGNN+GATGPT (Separately)& 23.45&15.94	&37.62	\\
&MTGNN+GATGPT (Together)& 24.86&16.75	&40.37	\\
&MTGNN (Separately) & 25.07&17.14	&40.54 \\
&MTGNN (Together)   & 26.58&	18.15&42.07 \\
&MTGNN (Complete)  &32.65 &	23.46&49.54 \\
\toprule

\end{tabular}}
\end{table}

\subsection{Efficiency} \label{effi}

\begin{figure}
\center
\includegraphics[width=\linewidth]{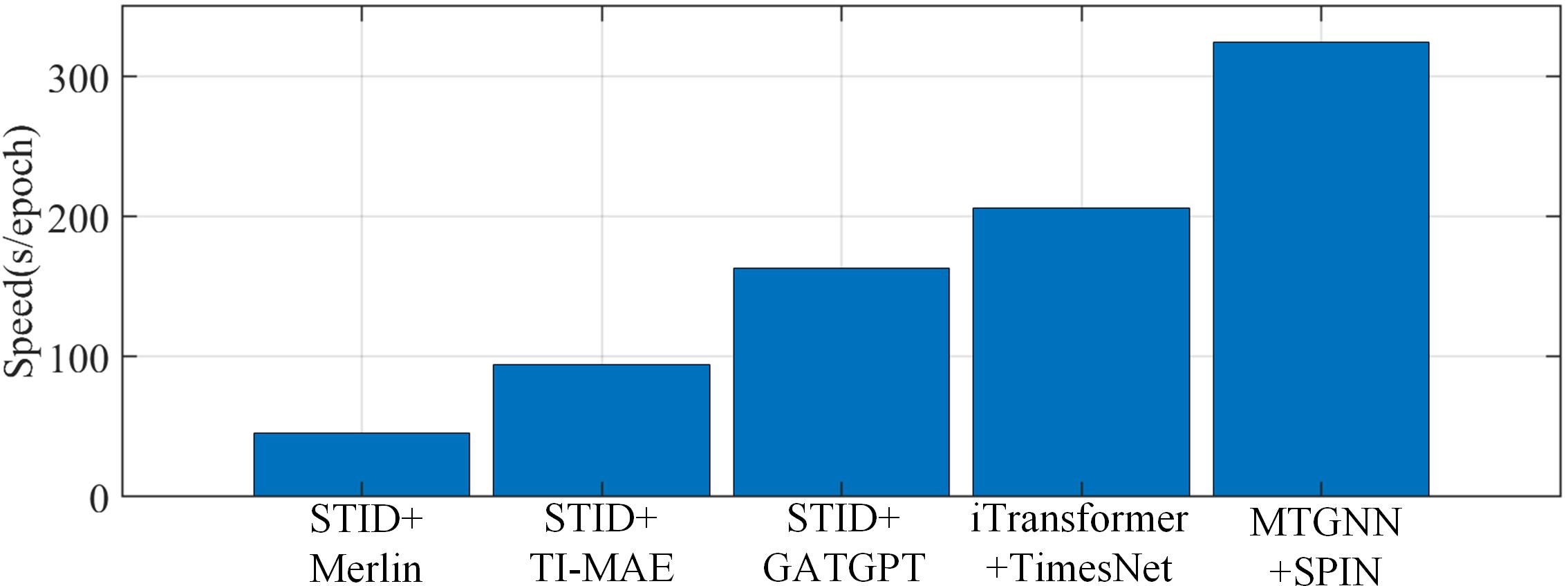}
\caption{Training time for each epoch of several models.}
\label{fig8}
\end{figure}

We compare the training times on the PEMS04 dataset for STID + Merlin, STID+GPT2, STID+GATGPT, iTransformer+TimesNet, and MTGNN+SPIN. Considering that STID+Merlin only needs to be trained once to adapt to different missing rates, whereas the other baselines require separate training sessions for each missing rate, we directly record the training time of the proposed STID+Merlin for a single epoch and sum up the training times for each missing rate for the other baselines. The experimental equipment is the 8 $\times$ RTX 4090 graphics card. Figure \ref{fig8} displays the average training time per epoch for these models. Despite incorporating components such as contrastive learning and knowledge distillation during the training process, STID+Merlin also achieves satisfactory results in terms of efficiency. Besides, Since neither the imputation model nor the teacher model is needed during the inference phase, STID+Merlin offers greater efficiency advantages during inference.

\section{Conclusion}

This paper aims to enhance the robustness of existing forecasting models on MTSF with unfixed missing rates, which involves two main challenges: (1) missing values disrupting the semantics (global and local information) of MTS; (2) the missing rate is unfixed at different time periods in the real world. To this end, we propose Merlin based on offline knowledge distillation and multi-view contrastive learning. Merlin can assist existing models in effectively achieving semantic alignment between incomplete observations with different missing rates and complete observations, thereby significantly enhancing their robustness to the unfixed missing rate. Extensive experiments show that the proposed model can achieve satisfactory forecasting results on all datasets and settings. Additionally, with the help of Merlin, existing forecasting models only need to be trained once to adapt to unfixed missing rates. In future work, we plan to investigate the effects of offline knowledge distillation when the teacher and student models utilize different neural network structures.

\begin{acks}
This work is supported by NSFC No. 62372430,  No. 62476264, and No. 62406312; the Postdoctoral Fellowship Program of CPSF under Grant Number GZC20241758 and BX20240385; the Youth Innovation Promotion Association CAS No.2023112; and HUA-Innovation fundings.
\end{acks}

\bibliographystyle{ACM-Reference-Format}
\balance
\bibliography{sample-base}

\appendix

\section{Compared with Different Loss Functions}
In terms of constructing the loss function, this paper uses L1 Loss to evaluate the difference between the forecasting results of the student model and the ground truth. In addition, L2 Loss is used to evaluate the difference between the student model and the teacher model. To better analyze the impact of the loss function on the results, we consider using only one of the loss functions or swapping the use of the above two loss functions. Besides, considering that the KL-divergence is also commonly used to evaluate the similarity between different distributions, we use the KL divergence as a new loss function to replace hidden representation distillation and conducted experiments. Table \ref{tab7} shows the MAE values of Merlin and other loss functions. The experimental results show that Merlin can get the best result. Additionally, compared to KL divergence, the MSE loss achieves better results. The main reason is that KL divergence focuses on improving the similarity between the distributions of representations, while MSE focuses on minimizing the numerical differences between representations. In summary, MTSF with missing values is a regression task, where minimizing numerical differences is more important.

\begin{table} [h]
\centering
\caption{MAE values of the proposed method and other loss functions (The best results are shown in \textbf{bold}).}\label{tab7}
\scalebox{0.95}{\begin{tabular}{cccccc}
\toprule
\multirow{2}*{Datasets} & \multirow{2}*{Methods} &\multicolumn{4}{c}{Missing rates}\\

\cline{3-6} 
& &25\% &50\% &75\% &90\%	\\

\midrule

\multirow{5}*{METR-LA} 
&Proposed&	\textbf{3.21}&\textbf{3.28}	&\textbf{3.43}	&\textbf{3.65}\\
&L1& 3.23&3.31 &3.44 &3.67\\
&L2& 3.28&3.37 &3.53 &3.73\\
&KL-divergence& 3.26&3.34&3.49&3.71\\
&Swapping& 3.27&3.35 &3.51 &3.72\\

\midrule

\multirow{5}*{Global Wind} 
&Proposed	&\textbf{5.76}&\textbf{5.85}	&\textbf{5.97}	&\textbf{6.08}\\
&L1 &5.79&5.87 &6.01 &6.10\\
&L2 & 5.81& 5.92& 6.06&6.15\\
&KL-divergence&5.80 &5.89&6.02&6.11\\
&Swapping & 5.83&5.91&6.04&6.17\\

\toprule

\end{tabular}}
\end{table}

\section{Experiments When the Performance of the Teacher Model is Degraded}\label{teacher}

Both existing imputation models and our model assume that complete data can be collected during the training phase \citep{chen2024laplacian}. However, considering the possibility of incomplete data collection in real-world scenarios (i.e., missing values in the training set), the teacher model and imputation models might be trained on multivariate time series with missing values, potentially leading to degraded performance. Therefore, it is crucial to evaluate the effectiveness of Merlin under such conditions. In this section, we simulate scenarios where the training data for the teacher model has the 5\% missing rate (this challenge also faced by imputation methods) and evaluate the improvements brought by Merlin and other imputation methods to different backbones under these conditions. Specifically, the original data is first modified to simulate the 5\% missing rate. Subsequently, these datasets are further processed to simulate higher missing rates of 25\%, 50\%, 75\%, and 90\%. Merlin and baselines is trained on datasets with the 5\% missing rate, as well as the datasets with subsequent missing rates of 25\%, 50\%, 75\%, and 90\%. 

Table \ref{tab13} shows the MAE values of Merlin and other methods. The following conclusions can be drawn: (1) Even when the data quality of the training sets for the teacher model decreases, Merlin can still effectively enhance the forecasting performance of several backbone models.
(2) Compared to imputation models, Merlin demonstrates superior capability in restoring the forecasting performance of different backbones, further proving its practical value in real-world scenarios.

\begin{table}
\centering
\caption{MAE values of Merlin and other methods (The missing rate of the training set is 5\%).}\label{tab13}
\scalebox{0.8}{\begin{tabular}{cc|cccc|cccc}
\toprule

\multirow{2}*{Backbone} & \multirow{2}*{Methods}&\multicolumn{4}{c|}{METR-LA}&\multicolumn{4}{c}{Global Wind}\\

\cline {3-10}
& &25\% &50\% &75\% &90\% &25\% &50\% &75\% &90\%\\

\midrule
\multirow{6}*{\rotatebox{90}{STID}}&+Merlin	
&\textbf{3.25}	&\textbf{3.33}	&\textbf{3.46}	&\textbf{3.69}
&\textbf{5.80}	&\textbf{5.88}	&\textbf{6.01}&\textbf{6.11}\\
&+GATGPT
&3.31&3.43 & 3.56 &3.77
&5.86& 5.94&  6.11&6.17\\
&+GPT2	
&3.34&3.48&3.59&3.80 
&5.88&5.97&6.16&6.24\\
&+TI-MAE	
&3.38&3.50&3.62&3.82
&5.90&6.01	&6.14	&6.21\\
&+SPIN	
&3.30&3.49&3.64&3.83   
&5.85&5.98	&6.15	&6.26\\
&raw	
&3.44& 3.57&  3.79 &3.96
&5.95&6.09&6.18  & 6.30\\
\midrule 

\multirow{6}*{\rotatebox{90}{DUET}}&+Merlin	
&\textbf{3.30}&\textbf{3.41} &\textbf{3.57} &\textbf{3.80}
&\textbf{5.85}&\textbf{5.95}	&\textbf{6.10}	&\textbf{6.17}\\
&+GATGPT
&3.32&3.48&3.60&3.86
&5.87&5.98&6.12&6.20\\
&+GPT2	
&3.35&3.52&3.64&3.91
&5.91&6.01	&6.14	&6.24\\
&+TI-MAE	
&3.35&3.49&3.59&3.86
&5.90&6.02&6.15&6.25\\
&+SPIN	
&3.29&3.51&3.64&3.91
&5.87&5.99&6.16&6.27\\
&raw	
&3.45&3.62&3.84&4.01
&5.97&6.11&6.19&6.33\\
\midrule 
\multirow{6}*{\rotatebox{90}{TimeMixer}}&+Merlin
&\textbf{3.35}&\textbf{3.45} &\textbf{3.61} &\textbf{3.84}
&\textbf{5.86}&\textbf{5.98} &\textbf{6.11} &\textbf{6.21}\\
&+GATGPT
&3.39&3.52&3.67&3.90
&5.90&6.03    &6.15  &6.24\\
&+GPT2	
&3.42&3.57&3.70&3.93
&5.91&6.06&6.17&6.26\\
&+TI-MAE	
&3.45&3.59&3.74&3.98
&5.94&6.08&6.16&6.27\\
&+SPIN	
&3.41&3.60&3.78&3.96
&5.88&6.09&6.18&6.30\\
&raw	
&3.47&3.63&3.87&4.03
&5.97&6.12&6.21&6.35\\
\midrule
\multirow{6}*{\rotatebox{90}{MTGNN}}&+Merlin
&\textbf{3.27}&\textbf{3.39} &\textbf{3.54} &\textbf{3.77}
&\textbf{5.84}&\textbf{5.97} &\textbf{6.07} &\textbf{6.15}\\
&+GATGPT
&3.30&3.42&3.57&3.79
&5.88&6.01&6.11&6.19\\
&+GPT2	
&3.34&3.47&3.64&3.89
&5.91&6.03&6.15&6.23\\
&+TI-MAE	
&3.37&3.49&3.67&3.91
&5.92&6.04&6.16&6.25\\
&+SPIN	
&3.31&3.45&3.59&3.84
&5.87&5.98&6.13&6.22\\
&raw	
&3.43&3.58&3.81&3.95
&5.96&6.10&6.20&6.34\\

\toprule

\end{tabular}}
\end{table}

\section{Compared with Multi-Stage Training}
Considering that different training processes can affect the overall performance of the model, this section compares the effects of multi-stage training with adding all loss functions. The multi-stage training strategy used to construct the comparative experiment includes the following two aspects \citep{mukherjee2020xtremedistil}: (1) Three-stage training: Firstly, train the model using the loss function of contrastive learning, then optimize the student model using the loss function of knowledge distillation, and finally optimize the student model using the L1 loss. (2) Two-stage training: Firstly, train the model using contrastive learning. Then optimize the student model using the combination of knowledge distillation and the L1 loss. 

Table \ref{tab6} shows the RMSE values of the proposed method and other multi-stage training methods. Experimental results show that compared with the multi-stage training strategy, the proposed method can achieve better forecasting results. The main reason is the problem of information forgetting in multi-stage training, which limits the performance of STID. 

\begin{table}[h]
\centering
\caption{RMSE values of the proposed method and other multi-stage training methods  (The best results are shown in \textbf{bold}).}\label{tab6}
\scalebox{0.95}{\begin{tabular}{cccccc}
\toprule
\multirow{2}*{Datasets} & \multirow{2}*{Methods} &\multicolumn{4}{c}{Missing rates}\\

\cline{3-6} 
& &25\% &50\% &75\% &90\%	\\

\midrule

\multirow{3}*{METR-LA} 
&Proposed	&\textbf{6.64}&\textbf{6.78} 	&\textbf{7.18}	&\textbf{7.64}\\
&Two-stage	&6.67&6.83	&7.25	&7.67\\
&Three-stage	&6.71&6.91	&7.27	&7.72\\

\midrule

\multirow{3}*{PEMS04} 
&Proposed	&\textbf{31.65}&\textbf{32.37}	&\textbf{34.45}	&\textbf{37.07}\\
&Two-stage	&32.08&32.94	&34.56	&37.44\\
&Three-stage	&32.34&33.06	&34.82	&37.61\\

\toprule

\end{tabular}}
\end{table}

\section{Notations}\label{note}

Some commonly used notations are presented in Table \ref{tab1}.

\begin{table}[h]
\small
\centering
  \caption{Frequently used notation.}
  \label{tab1}
\begin{tabular}{ccc}
\toprule
    Notation&size&Definitions\\
\midrule
$N_H$& Constant&Length of historical observations\\
$N_L$& Constant&Length of forecasting results\\
$N_s$&Constant&Batch size\\
$N_v$&Constant&Number of variables\\
$N_c$&Constant&Number of features\\
$m$&Constant&Number of missing rates\\
$X$&$N_v \times N_H\times N_c$& Complete observations\\
$X_M$&$N_v\times N_H\times N_c$&incomplete observations\\
$Y$&$N_v\times N_L$&Forecasting results\\
$\text{FC}$&Functions&Fully connected layer\\
\text{ReLU}&Functions&Activation function ReLU\\
Mean &Functions& The mean of the Tensor\\
\text{softmax}&Functions&Activation function softmax\\
\toprule
\end{tabular}
\end{table}

\end{document}